\documentclass{article}
\usepackage{spconf,amsmath,graphicx, hyperref}
\usepackage{float}
\usepackage{amssymb}

\title{RT-X Net: RGB-Thermal cross attention network for Low-Light Image Enhancement}
\name{Raman Jha$^1$, Adithya Lenka$^1$, Mani Ramanagopal$^2$, Aswin Sankaranarayanan$^2$, Kaushik Mitra$^1$}
\address{$^1$Indian Institute of Technology, Madras, $^2$Carnegie Mellon University}
\begin{document}
%
\maketitle
\begin{abstract}
 In nighttime conditions, high noise levels and bright illumination sources degrade image quality, making low-light image enhancement challenging. Thermal images provide complementary information, offering richer textures and structural details. We propose RT-X Net, a cross-attention network that fuses RGB and thermal images for nighttime image enhancement. We leverage self-attention networks for feature extraction and a cross-attention mechanism for fusion to effectively integrate information from both modalities. To support research in this domain, we introduce the Visible-Thermal Image Enhancement Evaluation (V-TIEE) dataset, comprising 50 co-located visible and thermal images captured under diverse nighttime conditions. Extensive evaluations on the publicly available LLVIP dataset and our V-TIEE dataset demonstrate that RT-X Net outperforms state-of-the-art methods in low-light image enhancement. The code and the V-TIEE can be found here https://github.com/jhakrraman/rt-xnet. 
\end{abstract}
\section{Introduction}
\label{sec:intro}


Image enhancement is crucial in various domains, such as medical imaging, surveillance, and environmental monitoring, as it improves image quality for more accurate information and better decision-making outcomes. Building on the success of convolutional neural networks (CNNs) in image denoising \cite{lyu2020degan}, CNN-based architectures have also been applied to deblurring \cite{nah2017deep, kupyn2018deblurgan} and super-resolution \cite{zhang2018learning}. Recent low-light image enhancement networks primary rely on deep learning models inspired by Retinex theory \cite{wei2018deep, cai2023retinexformer}, SNR-aware transformers \cite{xu2022snr}, or diffusion models \cite{hou2024global}. Notably, none of these approaches have explored the potential benefits of incorporating features from thermal images, which provide crucial information at night and in bad weather conditions. 


\begin{figure}
    \centering
    \includegraphics[height= 4.5cm, width=\linewidth]{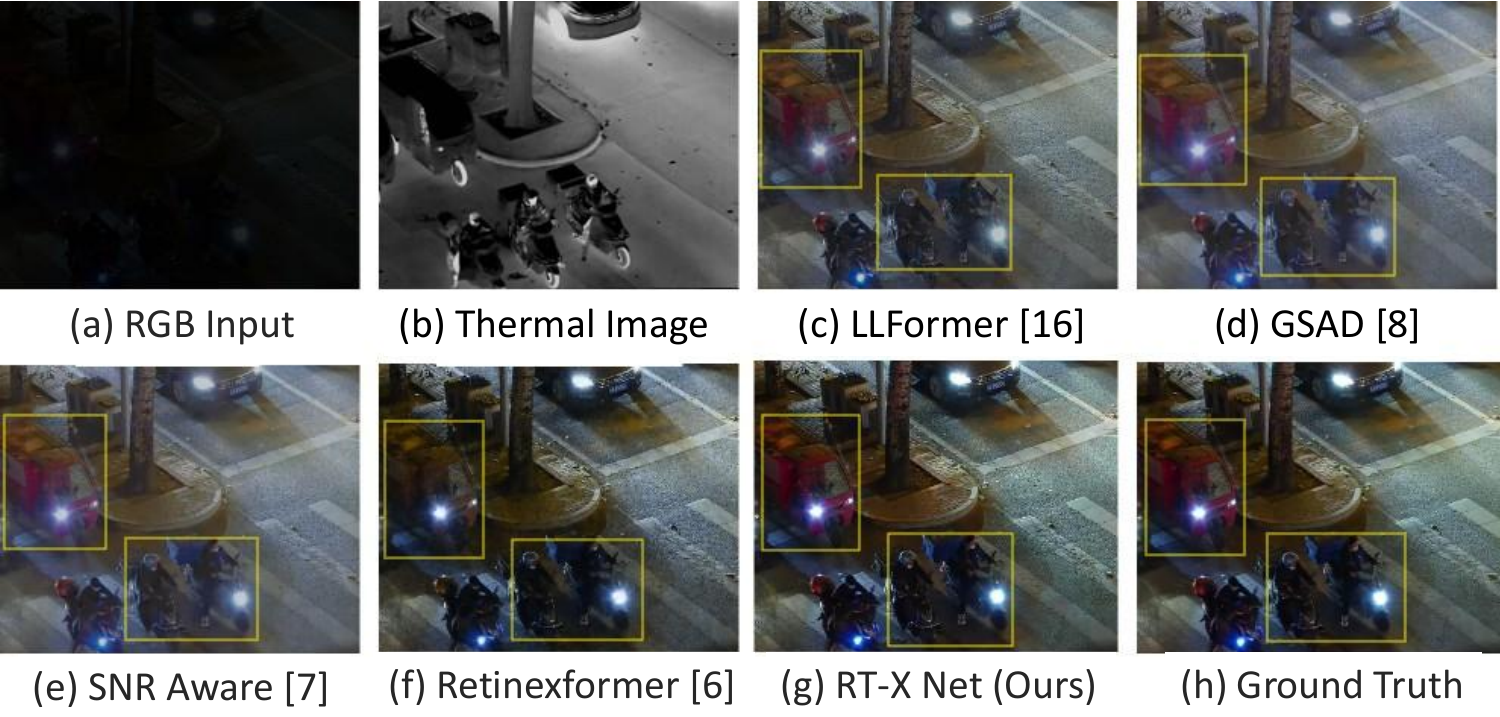}
    \vspace{-0.3 cm}
    \caption{Visual comparison on an extremely low light image. Our method achieves better enhancement compared to recent state-of-the-art methods.}
    \label{fig:overview}
    \vspace{-0.3 cm}
\end{figure}

The utilization of thermal imagery holds notable advantages in scenarios characterized by low-light conditions, owing to its consistent information capture regardless of illumination and light variations. Recent research \cite{farooq2023role} investigating the influence of thermal images in autonomous driving has shown broader implications, highlighting diverse aspects of thermal imagery utilization. Previously, researchers \cite{ha2017mfnet} have explored RGB-thermal fusion for tasks such as image segmentation, particularly in autonomous driving applications. More recent endeavors \cite{zhang2023cmx} have delved into cross-modality fusion, leveraging RGB and thermal images to generalize fusion techniques across multiple modalities, particularly emphasizing semantic segmentation tasks.

Cross-attention mechanisms are well-established in the field; however, their application to RGB-thermal fusion for low-light image enhancement remains a novel and unexplored approach. To the best of our knowledge, no prior research has utilized a cross-attention network for this purpose. As shown in Fig.\ref{fig:overview}, our method significantly outperforms existing techniques, such as the direct concatenation of RGB and thermal modalities. The key contributions of our work are summarized as follows:


\begin{itemize}

\item We propose \textit{RT-X Net}, a transformer-based cross-attention network that enhances low-light images by integrating thermal information, improving image quality, textures, and edges under extreme lighting conditions.  

\item We introduce a Cross-Attention Module that fuses RGB and thermal features, dynamically integrating illumination into the attention mechanism for context-aware fusion and adaptability to challenging lighting conditions.  

\item We present the real-world Visible-Thermal Image Enhancement Evaluation (V-TIEE) dataset with 50 diverse indoor and outdoor scenes. Experiments show RT-X Net, trained on LLVIP, generalizes to V-TIEE without fine-tuning.  

\end{itemize}

  %

\section{Previous work}
\label{sec:related_works}

\subsection{Low-Light Image Enhancement}


Low-light image enhancement is challenging due to dynamic illumination and noise in nighttime conditions. Traditional methods like histogram stretching \cite{celik2011contextual} had limited success. In contrast, CNN-based approaches \cite{chen2018learning} have improved results leveraging the retinex theory \cite{wei2018deep, cai2023retinexformer} to use illumination maps and reflectance for enhancement. Transformer-based methods \cite{wang2022low, wang2023ultra} have also emerged, utilizing self-attention networks for efficient feature extraction and improved enhancement. Xu et al. \cite{xu2022snr} proposed SNR-aware transformers for spatial-varying low-light image enhancement. Other methods, such as the Global Structure-Aware Diffusion process \cite{hou2024global} and GAN-based approaches \cite{jiang2021enlightengan, choudhary2023elegan}, have also improved performance in low light conditions. However, existing methods struggle with extremely low-light images. Shi et al. \cite{shi2023rgb} used cross-attention networks with RGB and LUT features but focused on well-exposed images. To overcome these limitations, we propose a cross-attention network that uses thermal images to enhance RGB quality in dark scenes.

\subsection{Thermal Image Fusion}
Thermal imaging captures valuable information regardless of noise and illumination. Recent studies have explored fusing thermal and RGB images, such as Ha et al.'s MFNet \cite{ha2017mfnet}, a CNN architecture for multi-spectral image segmentation, and Sun et al.'s RTFNet \cite{sun2019rtfnet}, which uses ResNet for RGB-thermal fusion in semantic segmentation. Zhang et al. \cite{zhang2023cmx} proposed CMX, a transformer-based framework for cross-modal fusion in RGB-X segmentation, showing thermal fusion's versatility across modalities. Cao et al. \cite{cao2023deep} used CNNs for thermal integration in low-light image enhancement. Transformers have already surpassed CNNs using only RGB. We enhance low-light images by integrating RGB and thermal data with a cross-attention transformer, leveraging thermal features for extreme low-light conditions.


\begin{figure*}[h]
    \centering
    \includegraphics[height = 7cm, width=\textwidth]{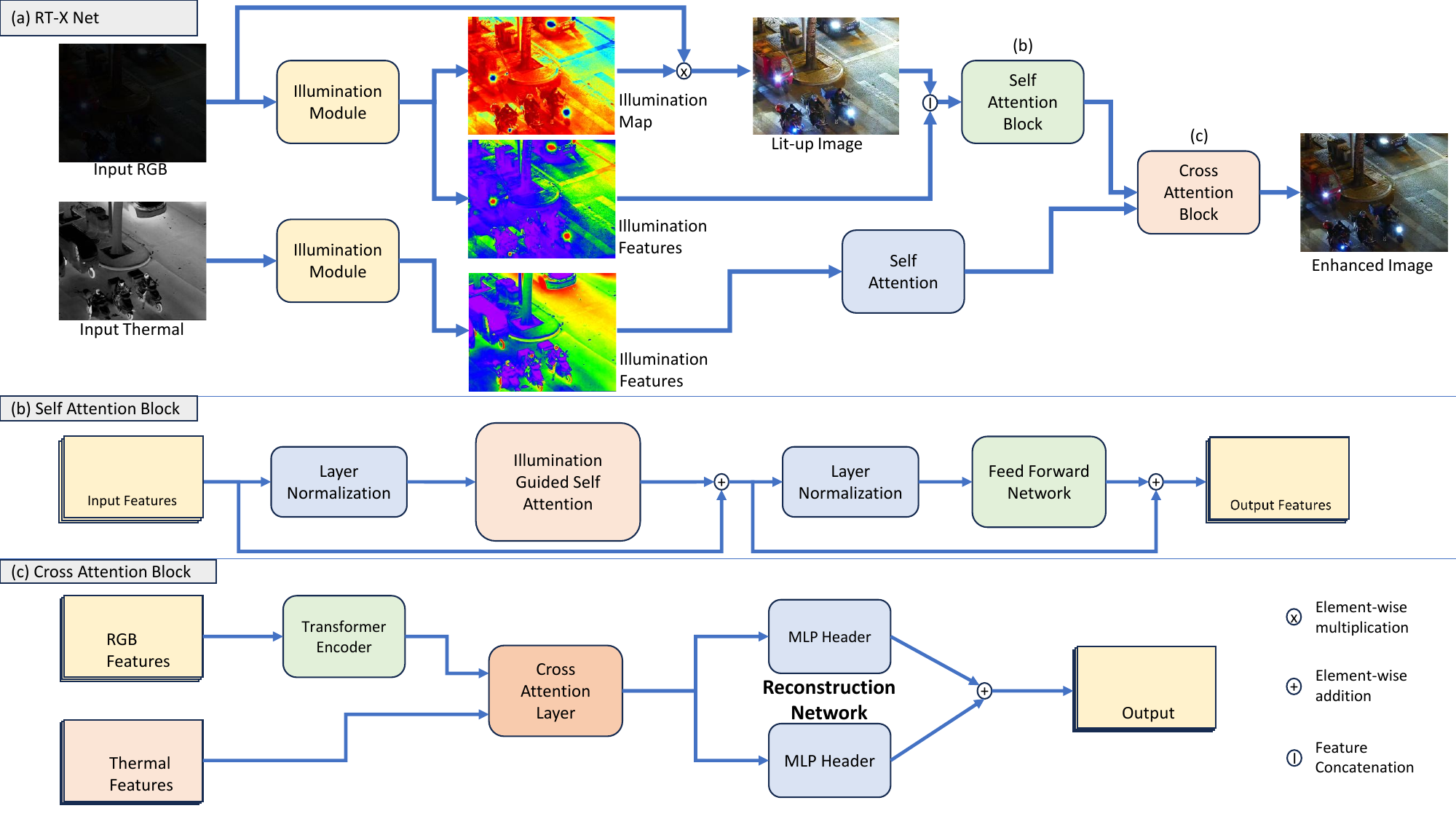}
    \caption{Our method overview: (a) In RT-X Net, RGB and thermal images are input to an illumination estimator to extract image features, which undergo self-attention and are fused via a cross-attention network to produce enhanced low-light images. (b) Workflow of the self-attention network. (c) Cross-attention network architecture for fusing RGB and thermal features.}
    \label{fig:model}
\end{figure*}

\section{Proposed Method}
\label{sec:methods}



\subsection{Retinex-based Decomposition and Illumination Guidance}
According to Retinex theory \cite{wei2018deep, cai2023retinexformer}, a low-light image $\mathbf{I} \in \mathbf{{R}^{H \times W \times 3}}$ can be expressed as an element-wise product of a \emph{reflectance} image $\mathbf{R}$ (same dimensions as $\mathbf{I}$) and an \emph{illumination} map $\mathbf{M} \in \mathbb{R}^{H \times W}$:
\begin{equation}
    \mathbf{I} = \mathbf{R} \odot \mathbf{M}.
\end{equation}
In low-light conditions, $\mathbf{M}$ captures lighting variations, while $\mathbf{R}$ preserves scene structure. Following \cite{cai2023retinexformer}, our illumination estimator predicts both $\mathbf{M}$ and \emph{illumination features} $\mathbf{F}_{\mathrm{illum}}$ from input RGB and thermal images. Formally,
\begin{equation}
    \mathbf{F}_{\mathrm{illum}},\ \mathbf{M} \;=\; \text{IlluminationEstimator}\bigl(\mathbf{I}\bigr),
\end{equation}
where $\mathbf{I}$ is either the RGB or thermal image. For the RGB branch, we multiply the input image by the predicted map $\mathbf{M}$ to create a partially ``illuminated’’ version before self-attention. For thermal branch, we only use $\mathbf{F}_{\mathrm{illum}}$, as thermal captures scene details without visible-light intensity.

\subsection{Self-Attention for Feature Extraction}
To encode spatial context and long-range dependencies in low-light images, we adopt a self-attention network inspired by transformers \cite{wang2022low, wang2023ultra}. Let $\mathbf{x}^{l} \in \mathbb{R}^{H \times W \times C}$ be the feature input, and let $\mathbf{x}^{s}$ be the self-attended output:
\begin{equation}
    \mathbf{x}^{s} = \mathrm{SelfAttn}\bigl(\mathbf{x}^{l}, \mathbf{F}_{\mathrm{illum}}\bigr),
\end{equation}
where $\mathrm{SelfAttn}(\cdot)$ projects queries, keys, and values in order to capture salient features. Concretely, queries and keys are normalized dot products:
\begin{equation}
    \mathrm{Attn}(\mathbf{Q},\mathbf{K},\mathbf{V}) \;=\; 
     \mathrm{softmax}\!\Bigl(\frac{\mathbf{Q}\,\mathbf{K}^{\mathsf{T}}}{\sqrt{d_k}}\Bigr) \mathbf{V},
\end{equation}
and include illumination features as a learnable re-weighting term on the values $\mathbf{V}$ so that bright or crucial regions are emphasized even when overall image brightness is low. This yields two separate self-attended features: $\mathbf{x}^{a}_{\mathrm{RGB}}$ and $\mathbf{x}^{a}_{\mathrm{Therm}}$.

\subsection{Cross-Attention for RGB--Thermal Fusion}
After extracting self-attended representations from both RGB and thermal images, we fuse them via a cross-attention module. Let $\mathbf{x}^{a}_{\mathrm{RGB}}$ and $\mathbf{x}^{a}_{\mathrm{Therm}}$ be the respective features:
\begin{equation}
    \label{eq:crossattn}
    \mathbf{x}^{c} = \mathrm{MCA}\bigl(\mathbf{x}^{a}_{\mathrm{RGB}} \,\|\, \mathbf{x}^{a}_{\mathrm{Therm}}\bigr),
\end{equation}
where $\mathrm{MCA}(\cdot)$ is a multi-head cross-attention operator. Queries are derived from one modality, while keys and values come from the other, thus allowing thermal-induced feature maps to refine and guide the RGB feature space. Because thermal images are invariant to darkness and glare, $\mathbf{x}^{a}_{\mathrm{Therm}}$ often provides sharper structure cues for extremely low-light scenes. After combining features from RGB and thermal images, a principle component analysis step is applied to reduce the channel dimensionality. This helps in controlling the computational overhead and guide the network to the most discriminative blend of RGB–thermal representations.

\subsection{Enhanced Image Reconstruction}

Finally, the fused feature $\mathbf{x}^{c}$ is passed into a reconstruction network, which consists of trainable MLP header mappings to generate the enhanced image as shown in Fig. \ref{fig:model}. During training, a Mean Absolute Error (MAE) loss function is minimized:
 
\begin{equation}
    \mathcal{L} \;=\; \frac{1}{N} \sum_{i=1}^N \bigl\|\hat{\mathbf{I}}_{\text{out}}^{(i)} \;-\; \mathbf{I}_{\mathrm{gt}}^{(i)}\bigr\|_1,
\end{equation}
where $\mathbf{I}_{\mathrm{gt}}^{(i)}$ represents ground-truth well-lit images. It finally produces an enhanced image from low-light image.
\section{Experiments}
\label{sec:expts}


\subsection{Real-world Low-light V-TIEE dataset}
We introduce the V-TIEE dataset, captured using a co-located visible (1440×1080px, 34° FOV) and thermal (640×512px, 24° HFOV) camera co-located using a gold dichroic mirror that transmits visible light and reflects thermal infrared. Final spatial alignment is refined using homography.

\begin{figure*}[ht]
    \centering
    \includegraphics[height = 3.5cm, width = 14cm]
    {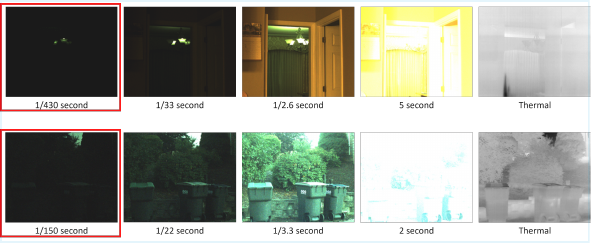}
    \caption{Multiple high-gain exposures (above) and low-gain exposures (below) were captured for indoor and outdoor scenes with corresponding thermal images. The red box marks input images for testing, while the green box highlights reference images.}
    \label{fig:high_gain}
\end{figure*}

\begin{figure*}[ht]
    \centering
    \includegraphics[height = 3.5cm, width = 14cm]
    {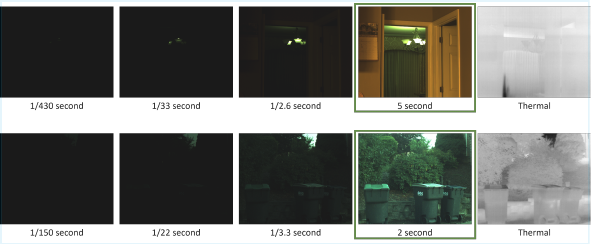}
    \label{fig:low_gain}
\end{figure*}

The dataset includes 50 indoor and outdoor scenes with RGB and thermal images under varying gain and exposure settings. Images were recorded at two gain levels with five exposure values each, while four scenes have ten exposures per gain. Gain ranges from $0 \textendash 44.99$ dB, and exposure times from $1/20000 \textendash 10$ seconds (Fig. \ref{fig:high_gain}). This dataset supports HDR research, low-light applications, and model generalization under realistic noise. More details are in the supplementary material. \url{https://sigport.org/sites/default/files/docs/Supplementary_11.pdf}.




\begin{figure*}[!ht]
    \centering
    \includegraphics[height = 15cm, width=\textwidth]{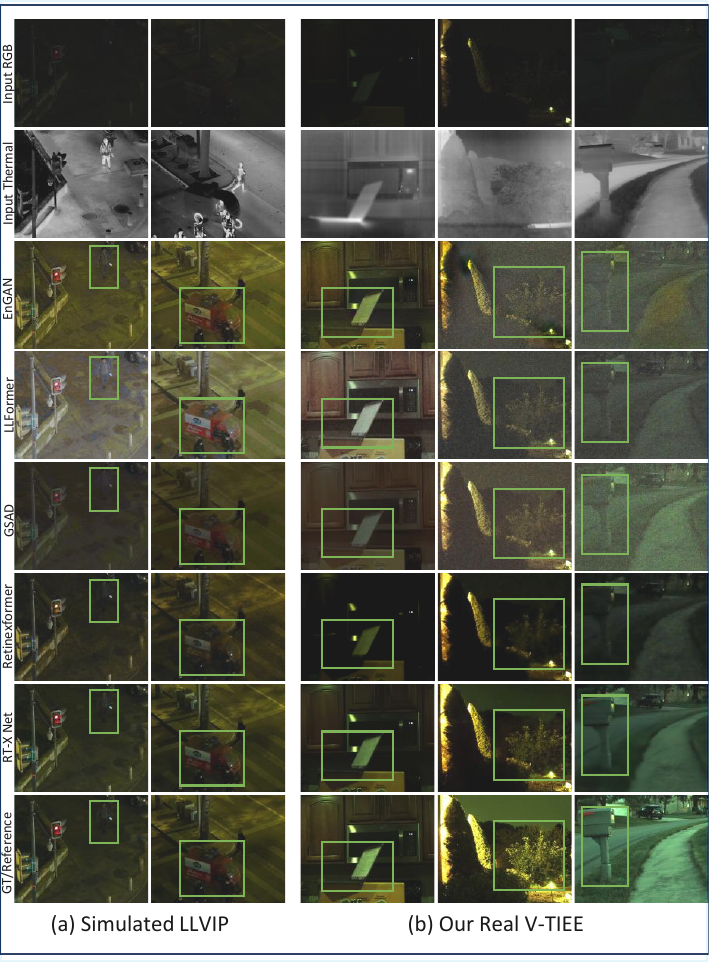}
    \caption{Qualitative results on the synthetic LLVIP dataset and real-world V-TIEE dataset. Columns denote different scenes. The first two rows show the input RGB and thermal images. The next five rows are the outputs from RT-X Net and state-of-the-art visible image enhancement algorithms. The last row shows the reference well-exposed image.}
    \label{fig:qualitative}
\end{figure*}

\subsection{Evaluation details}

\subsubsection{Datasets and Evaluation Metrics:}
We use the LLVIP dataset \cite{jia2021llvip}, selecting every 50th image for a diverse subset of 850 RGB-thermal pairs (805 train, 45 test). Low-exposure conditions are simulated by reducing exposure (×5–20) and adding noise \cite{hasinoff2010noise}. PSNR, and SSIM, are used for evaluation. We evaluate generalization on V-TIEE, using low-exposure images as input and high-exposure images as ground truth. Due to illumination differences with LLVIP, we use LPIPS and SSIM for perceptual and structural similarity.

\subsubsection{Implementation details:}

RT-X Net is implemented using the PyTorch framework. The model is trained with the Adam optimizer over  $1.5 \times 10$${^5}$ iterations. Training involves randomly cropping $128 \times 128$ patches from low-light images in the LLVIP dataset. A fixed learning rate of $2 \times 10$$^{-4}$ and a batch size of $4$ are used. Our model has 0.67 million parameters. For generalization and robustness, we employ random flipping and rotation as data augmentation. The Mean Absolute Error (MAE) is used as the loss function.


\subsection{Qualitative Analysis}
We evaluate RT-X Net against recent low-light enhancement methods using the LLVIP and V-TIEE datasets in Fig. \ref{fig:qualitative}. Cao et al. \cite{cao2023deep} proposed a CNN-based thermal-guided network; however, their results could not be compared due to the unavailability of their code. Prior methods, including GSAD \cite{hou2024global}, Retinexformer \cite{cai2023retinexformer}, LLFlow \cite{wang2022low}, KinD \cite{zhang2019kindling}, SNR Aware \cite{xu2022snr}, LLFormer \cite{wang2023ultra}, and EnGAN \cite{jiang2021enlightengan}, exhibited issues such as blur, artifacts, dark spots, color distortions, or exposure inconsistencies. RT-X Net surpasses prior limits by enhancing low-light images with sharp details, correct exposure, and minimal noise, delivering clear textures, pedestrians, and vehicles in LLVIP, as well as fine details like trees and mailboxes in V-TIEE.



\subsection{Quantitative Analysis}
To quantify the effectiveness, we compared RT-X Net to existing recent low-light image enhancement methods on two datasets: the publicly available LLVIP dataset and our real-world V-TIEE dataset in Table. \ref{tab:quantitative}. RT-X Net outperformed baseline methods in both datasets. Compared to the existing best-performing low-light image enhancement method, Retinexformer \cite{cai2023retinexformer}, RT-X Net achieves a $1.16$ dB gain on LLVIP and outperforms transformer-based SOTA methods. Using LPIPS and SSIM on V-TIEE, it shows strong quantitative results, confirming its effectiveness for low-light image enhancement.

\begin{table}[h]
\begin{center}
\begin{tabular}{|l|c|c|c|c|}
\hline
Method & \multicolumn{2}{c|}{LLVIP} & \multicolumn{2}{c|}{real-world V-TIEE} \\
    &   PSNR $\uparrow$ & SSIM $\uparrow$ & LPIPS $\downarrow$ & SSIM $\uparrow$ \\
\hline\hline
KinD \cite{zhang2019kindling} & 20.41 & 0.54  & 0.29  &  0.52 \\
LLFlow \cite{wang2022low} & 21.53 & 0.57  &  0.27 &  0.56 \\
LLFormer\cite{wang2023ultra} & 22.27 &  0.61 & 0.26  &  0.60 \\
SNR Aware \cite{xu2022snr} & 22.87 & 0.64  & 0.23  &  0.61 \\
GSAD Net \cite{hou2024global}& 24.43 &  0.69 &  0.21 &  0.63 \\
EnGAN \cite{jiang2021enlightengan} & 25.67 & 0.74  & 0.17  &  0.65 \\
Retinexformer \cite{cai2023retinexformer}& 26.59 & 0.79  & 0.14  & 0.66  \\
\textbf{RT-X Net [Ours]}& \textbf{27.75} &  \textbf{0.85} &  \textbf{0.12} &  \textbf{0.71} \\
\hline
\end{tabular}
\end{center}
\caption{Quantitative comparison of RT-X Net with baseline methods on LLVIP and V-TIEE dataset.}
\label{tab:quantitative}
\end{table}



\subsection{Ablation Study}
The performance evaluation of RT-X Net is divided into two parts in the Table. \ref{tab:llvip_ablation}:

\textbf{1. Self-Attention Network:} An ablation study shows a significant performance drop when the cross-attention network is replaced with self-attention using only RGB images (Table 2), emphasizing the critical role of cross-attention in RT-X Net for effective low-light image enhancement.

\begin{table}[h]
\begin{center}
\begin{tabular}{|l|c|c|}
\hline
Ablation & PSNR $\uparrow$ & SSIM $\uparrow$\\
\hline\hline
 Self-Attention(Only RGB) &   26.42 &   0.73\\
 Thermal channel concatenation & 27.15 &  0.80 \\
 Cross-Attention &  \textbf{27.75} &  \textbf{0.85} \\
\hline
\end{tabular}
\end{center}
\caption{Ablation study on the LLVIP dataset.}
\label{tab:llvip_ablation}
\end{table}

\textbf{2. Another channel to concatenate Thermal Images:} Thermal data is incorporated by concatenating thermal images with RGB images to create a $4$-channel input, but this approach shows limited performance gains. Adding a dedicated layer for thermal images provides marginal improvement but remains less effective than the cross-attention network. These results highlight the superiority of the cross-attention network in leveraging thermal information for efficient low-light image enhancement.
\section{Conclusion}
\label{sec:conclusion}

We present RT-X Net, a transformer network utilizing cross-attention to leverage thermal imagery for low-light enhancement under extreme conditions. The experimental results demonstrate that the proposed approach outperforms recent state-of-the-art methods by effectively incorporating thermal information. Ablation studies further validate its effectiveness. Evaluations on the LLVIP and V-TIEE datasets demonstrate reduced noise, fewer artifacts, and improved illumination and color. Future work will focus on reducing model complexity for video processing and optimizing real-time applications like autonomous driving and robotics.

\bibliographystyle{IEEEbib}
\bibliography{egbib}

\end{document}